\documentclass{article}
\usepackage[final]{colm2026_conference}
\usepackage{iftex}
\RequireXeTeX
\usepackage{fontspec}
\newfontfamily\thaifont{Norasi}[
  Path           = fonts/,
  Extension      = .otf,
  UprightFont    = *,
  BoldFont       = *-Bold,
  ItalicFont     = *-Italic,
  BoldItalicFont = *-BoldItalic,
  Script         = Thai,
  Scale          = MatchLowercase]

\usepackage{ucharclasses}
\makeatletter
\newcommand{\thaiSavedFamily}{\rmdefault}
\newcommand{\thaiEnter}{\xdef\thaiSavedFamily{\f@family}\thaifont
  \XeTeXlinebreaklocale "th"\XeTeXlinebreakskip=0pt plus 0.1pt\relax}
\newcommand{\thaiLeave}{\fontfamily{\thaiSavedFamily}\selectfont
  \XeTeXlinebreaklocale ""\XeTeXlinebreakskip=0pt\relax}
\makeatother
\setTransitionsFor{Thai}{\thaiEnter}{\thaiLeave}

\usepackage[table]{xcolor}
\usepackage{microtype}
\usepackage{hyperref}
\usepackage{url}
\usepackage{booktabs}

\usepackage{lineno}
\usepackage{amsmath}
\usepackage{float}
\usepackage{subcaption}
\usepackage{multicol}
\usepackage[bottom]{footmisc}
\usepackage{diagbox}
\usepackage{multirow, booktabs}
\usepackage{caption}
\usepackage{pifont}% http://ctan.org/pkg/pifont
\usepackage{graphicx}
\usepackage{tikz}
\usetikzlibrary{arrows.meta,calc,positioning}

\definecolor{figInk}{HTML}{29323A}
\definecolor{figMuted}{HTML}{68727D}
\definecolor{figBlue}{HTML}{3F51B5}
\definecolor{figBluePale}{HTML}{EEF1FB}
\definecolor{figAmber}{HTML}{C98116}
\definecolor{figAmberPale}{HTML}{FFF6E6}
\definecolor{figCoral}{HTML}{D95D50}
\definecolor{figTeal}{HTML}{21867A}
\definecolor{figTealPale}{HTML}{EAF6F3}

\newcommand\sbullet[1][.75]{\mathbin{\vcenter{\hbox{\scalebox{#1}{$\bullet$}}}}}

\usepackage[nobar]{wayupaxa}
\reportstamp{Technical Report}
\resources{%
  \resourcerow{Collection}{\href{https://huggingface.co/collections/wayu-ai/wayu-paxa-tts}{\texttt{Wayu-Paxa-TTS}}}
  \resourcerow{Model}{\href{https://huggingface.co/wayu-ai/wayu-paxa-tts-edge}{\texttt{Wayu-Paxa-TTS-Edge}}}
  \resourcerow{Evaluation}{\href{https://github.com/wayu-research/thai-tts-eval}{\texttt{Thai-TTS-Eval}}}}

\title{Building and Evaluating Fixed-Voice Thai TTS \\ from Synthetic Speech}

\author{Kunat Pipatanakul$^{1,2}$, Potsawee Manakul$^{1}$, Warit Sirichotedumrong$^{3}$ \\ Sittipong Sripaisarnmongkol$^{3}$, Pakorn Nathong$^{2}$, Phatrasek Jirabovonvisut$^{2}$ \\ \vspace{0.35em} {\normalfont\small\color{wayuCedar} $^{1}$Wayu Research \quad $^{2}$Paxa Labs \quad $^{3}$Typhoon } }

\begin{document}

\ifcolmsubmission
\linenumbers
\fi

\maketitle

\begin{abstract}
In low-resource settings, deploying TTS typically requires choosing between a large voice-cloning model with costly inference or a compact fixed-voice system that requires a speaker-specific corpus. We study a third route: using a large voice-cloning model as a programmable data source to turn a short voice reference (e.g., 15 seconds) into a compact fixed-voice student trained entirely on synthetic speech. This setting makes pipeline design consequential: teacher errors become training targets, while filtering failed generations can reduce coverage of difficult texts. Thai further introduces challenges from ambiguous word boundaries, lexical tone, names and loanwords, numeric verbalization, and Thai--English code-switching. We study how text preparation, synthetic generation, quality filtering, rejection sampling, and frontend choices affect the resulting student, and where teacher limitations remain. We evaluate CER, Challenge-Set Keyword Accuracy, Prosody Pause Accuracy, speaker similarity, and speaking rate. The resulting 82M-parameter model, \textbf{Wayu-Paxa-TTS-Edge}, enables on-device Thai TTS without reference audio. It achieves 68.2\% Challenge-Set Keyword Accuracy (85.5\% of Gemini 3.1) and 91.4\% pause precision, outperforming its OmniVoice teacher\abstractnote{\textcolor{red}{Because OmniVoice was trained on in-the-wild speech, these synthetic voices may incidentally resemble real individuals; we did not crawl speech data or intentionally clone any real person.}} (89.9\%) and reaching 94.8\% of Gemini 3.1. It also achieves the lowest pause-placement error and intra-word pause rates among the three systems, and 3.7\% and 1.1\% CER on Thai and English, respectively. We open-source the model and evaluation framework for Thai TTS development.

\end{abstract}

\section{Introduction}
\label{sec:introduction}

In low-resource settings, deploying TTS typically requires choosing between a large
generative voice-cloning model that uses reference audio and GPU inference, or a compact
fixed-voice TTS system trained on a licensed single-speaker corpus. Modern multilingual
TTS systems can reproduce a speaker from only a few seconds of reference
audio~\citep{zhang2025minimaxspeech,hu2026qwen3tts,bosonai2026higgstts3,zhu2026omnivoice}.
These systems rely on large generative backbones, including autoregressive language
models and diffusion language models, and often scale training to large multilingual
speech collections. Their scale provides broad zero-shot capabilities but 
can be unnecessarily expensive when an application needs only a 
single organization-specific voice, such as for an interactive voice response (IVR) system or a personal brand creator.
By contrast, TTS architectures such as VITS and StyleTTS2 support compact fixed-voice
deployment~\citep{pmlr-v139-kim21f,li2023styletts2}; our student uses the 82M-parameter
Kokoro backbone~\citep{hexgrad2025kokoro}. This second route simplifies deployment but
requires a speaker-specific corpus.

Inspired by knowledge distillation in modern LLM development~\citep{pipatanakul2024typhoon2}, we study a third route. We similarly use a large voice-cloning model to
generate speech from a short voice reference (e.g., 15 seconds) and train a compact
fixed-voice student.

Prior low-resource work uses synthetic target-language speech before adapting to
several hours of real target-speaker audio~\citep{joshi-garera-2023-rapid}. A recent
Thai-specific system takes a data-intensive route, constructing large speech and text
collections with explicit tone and pause annotations~\citep{geng-etal-2025-scaling}.
Our route requires no speaker-specific corpus beyond the short reference. Synthetic
data are unbounded in count, but their quality and coverage remain bounded by what the
stochastic teacher can produce and what the pipeline can select.

Thai makes this challenging because of ambiguous word boundaries, lexical tone,
irregular names and loanwords, informal spellings, numeric verbalization, and
Thai--English code-switching. Thai orthography does not mark word boundaries
consistently~\citep{chormai-etal-2020-syllable}, and prior Thai TTS work explicitly
models tone and pause placement~\citep{geng-etal-2025-scaling}. Code-switched inputs add
another ambiguity: the intended rendition is often Thai-accented English rather than
native English. These cases expose failures that sentence-level CER obscures. A sentence
can have low CER while mispronouncing one critical expression, and a transcript can be
correct even when the waveform contains a pause within a word or at an implausible
juncture.

The central question is therefore not whether synthetic speech can train a student, but
which pipeline components affect performance, how to evaluate their effects, and where
teacher limitations remain. Our pipeline covers text preparation, synthetic speech
generation with OmniVoice~\citep{zhu2026omnivoice}, quality filtering, rejection
sampling, and training a compact Kokoro student~\citep{hexgrad2025kokoro}. We
evaluate CER, Challenge-Set Keyword Accuracy, Prosody Pause Accuracy, speaker
similarity, and speaking rate.

Our final model, \textbf{Wayu-Paxa-TTS-Edge}, is an 82M-parameter fixed-voice
Thai--English TTS system that supports on-device inference. The model
achieves 68.2\% Challenge-Set Keyword Accuracy and 91.4\% pause precision on Thai, with
CERs of 3.7\% and 1.1\% on Thai and English, respectively. Teacher analysis shows that
best-of-$K$ teacher sampling reveals substantial headroom: exact accuracy reaches
87.9\% at $K=118$, 15.1 percentage points above the 72.8\% teacher baseline. This
gain identifies improved sampling and selection as a path toward stronger training targets;
the unresolved items are concentrated on expressions underrepresented in the teacher's
training corpus, making data coverage the next challenge.

Our contributions are:
\begin{itemize}
  \item an end-to-end recipe for converting a short voice reference into a
  quality-controlled Thai synthetic corpus and a compact fixed-voice student;
  \item an evaluation framework that separates sentence-level CER, targeted
  pronunciation correctness, pause placement, speaker similarity, and speaking rate;
  \item controlled evidence for the effects of pause filtering, rejection sampling,
  pretrained initialization, and frontend policy, including gains from frontend changes
  without acoustic-model retraining;
  \item a teacher-support analysis through best-of-$K$ sampling, together with an Isan
  adaptation study showing that a 15-second reference can transfer voice identity and
  dialect forms to the fixed-voice student.
\end{itemize}

\section{From a Zero-Shot Teacher to a Fixed-Voice Student}
\label{sec:method}

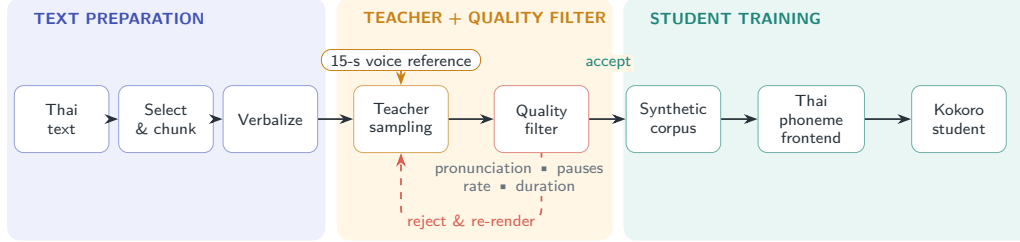
\begin{figure}[H]
  \centering
  \resizebox{\textwidth}{!}{\begin{tikzpicture}[
  x=1cm,
  y=1cm,
  font=\sffamily,
  >={Stealth[length=2.1mm,width=1.5mm]},
  flow/.style={->,draw=figInk,line width=0.75pt},
  stage/.style={draw=figMuted!45,fill=white,rounded corners=1.4mm,
    minimum height=1.02cm,text width=1.30cm,align=center,inner sep=2pt,
    font=\sffamily\scriptsize,text=figInk},
  badge/.style={circle,fill=figBlue,text=white,minimum size=6.8mm,inner sep=0pt,
    font=\sffamily\bfseries\footnotesize},
  pill/.style={draw=figAmber,fill=white,rounded corners=1.5mm,inner xsep=4pt,
    inner ysep=2pt,font=\sffamily\scriptsize,text=figInk}
]
  \path[use as bounding box] (0,0) rectangle (16,5.15);
  \fill[white] (0,0) rectangle (16,5.15);

  \node[badge] at (0.48,4.72) {2};
  \node[anchor=west,font=\sffamily\bfseries\large,text=figInk]
    at (0.92,4.72) {Build: zero-shot teacher to fixed-voice student};

  \fill[figBluePale,rounded corners=2mm] (0.20,0.45) rectangle (5.05,4.15);
  \fill[figAmberPale,rounded corners=2mm] (5.22,0.45) rectangle (9.42,4.15);
  \fill[figTealPale,rounded corners=2mm] (9.59,0.45) rectangle (15.80,4.15);

  \node[anchor=west,font=\sffamily\bfseries\scriptsize,text=figBlue]
    at (0.48,3.83) {TEXT PREPARATION};
  \node[anchor=west,font=\sffamily\bfseries\scriptsize,text=figAmber]
    at (5.50,3.83) {TEACHER + QUALITY FILTER};
  \node[anchor=west,font=\sffamily\bfseries\scriptsize,text=figTeal]
    at (9.87,3.83) {STUDENT TRAINING};

  \node[stage,draw=figBlue!55] (text) at (1.03,2.30) {Thai\\text};
  \node[stage,draw=figBlue!55] (chunk) at (2.62,2.30) {Select\\\& chunk};
  \node[stage,draw=figBlue!55] (verb) at (4.21,2.30) {Verbalize};

  \node[stage,draw=figAmber!70] (sample) at (6.20,2.30) {Teacher\\sampling};
  \node[stage,draw=figCoral!70] (filter) at (8.34,2.30) {Quality\\filter};
  \node[pill] (reference) at (6.20,3.20) {15-s voice reference};

  \node[stage,draw=figTeal!65] (corpus) at (10.35,2.30) {Synthetic\\corpus};
  \node[stage,draw=figTeal!65,text width=1.48cm] (frontend) at (12.45,2.30)
    {Thai phoneme\\frontend};
  \node[stage,draw=figTeal!65] (student) at (14.70,2.30) {Kokoro\\student};

  \draw[flow] (text.east) -- (chunk.west);
  \draw[flow] (chunk.east) -- (verb.west);
  \draw[flow] (verb.east) -- (sample.west);
  \draw[flow] (sample.east) -- (filter.west);
  \draw[flow] (filter.east) -- (corpus.west);
  \node[font=\sffamily\scriptsize,text=figTeal,fill=figAmberPale,inner sep=1pt]
    at (9.36,3.15) {accept};
  \draw[flow] (corpus.east) -- (frontend.west);
  \draw[flow] (frontend.east) -- (student.west);
  \draw[flow,draw=figAmber] (reference.south) -- (sample.north);

  \draw[->,draw=figCoral,line width=0.75pt,dashed,rounded corners=2mm]
    (filter.south) -- (8.34,0.73) -- (6.20,0.73) -- (sample.south);
  \node[align=center,font=\sffamily\scriptsize,text=figMuted,
    fill=figAmberPale,rounded corners=1mm,inner sep=2pt] at (8.00,1.42)
    {pronunciation \textbullet{} pauses\\rate \textbullet{} duration};
  \node[font=\sffamily\scriptsize,text=figCoral,fill=figAmberPale,inner sep=1pt]
    at (7.27,0.73) {reject \& re-render};
\end{tikzpicture}}
  \caption{Distilling a 15-second voice reference into fixed-voice Thai TTS. A
  zero-shot teacher renders prepared Thai text; quality filtering retains suitable
  candidates and triggers re-rendering of eligible failures, and the resulting corpus
  trains a Kokoro student with a Thai--English phoneme frontend.}
  \label{fig:pipeline}
\end{figure}

We distill a zero-shot voice-cloning teacher into a fixed-voice student through the
three-stage pipeline shown in Figure~\ref{fig:pipeline}: (1) Thai text preparation and
verbalization, (2) teacher sampling and quality filtering, and (3) student training. Given
Thai text and one of 12 frozen OmniVoice-designed voice references%
\footnote{\url{https://huggingface.co/k2-fsa/OmniVoice}}~\citep{zhu2026omnivoice}, the teacher generates
candidate utterances for evaluation by a quality filter. The resulting utterances form
the synthetic corpus used to train the Kokoro%
\footnote{\url{https://huggingface.co/hexgrad/Kokoro-82M}} student~\citep{hexgrad2025kokoro,li2023styletts2}.

\begin{figure}[H]
  \centering
  \resizebox{\textwidth}{!}{\begin{tikzpicture}[
  x=1cm,
  y=1cm,
  font=\sffamily,
  >={Stealth[length=2.1mm,width=1.5mm]},
  metric/.style={fill=white,rounded corners=1.5mm,minimum width=3.05cm,
    minimum height=1.86cm,align=center,inner sep=3pt,line width=0.75pt},
  badge/.style={circle,fill=figTeal,text=white,minimum size=6.8mm,inner sep=0pt,
    font=\sffamily\bfseries\footnotesize},
  input/.style={draw=figTeal,fill=figTealPale,rounded corners=3mm,
    minimum width=1.95cm,minimum height=0.62cm,align=center,
    font=\sffamily\bfseries\scriptsize,text=figTeal}
]
  \path[use as bounding box] (0,0) rectangle (16,4.15);
  \fill[white] (0,0) rectangle (16,4.15);

  \node[badge] at (0.48,3.72) {3};
  \node[anchor=west,font=\sffamily\bfseries\large,text=figInk]
    at (0.92,3.72) {Evaluate: complementary evidence beyond CER};

  \node[input] (speech) at (1.28,2.50) {Synthesized\\speech};
  \coordinate (busstart) at (2.55,2.50);
  \coordinate (busend) at (14.40,2.50);
  \draw[->,draw=figTeal,line width=0.9pt] (speech.east) -- (busstart);
  \draw[draw=figTeal,line width=0.9pt] (busstart) -- (busend);

  \node[metric,draw=figBlue] (correct) at (3.75,1.15) {};
  \node[metric,draw=figCoral] (pause) at (7.30,1.15) {};
  \node[metric,draw=figTeal] (voice) at (10.85,1.15) {};
  \node[metric,draw=figAmber] (timing) at (14.40,1.15) {};

  \foreach \target in {correct,pause,voice,timing}
    \draw[->,draw=figTeal,line width=0.75pt] (\target.north |- busstart) -- (\target.north);

  \node[font=\sffamily\bfseries\scriptsize,text=figBlue]
    at ([yshift=-0.28cm]correct.north) {Correctness};
  \draw[draw=figBlue!25,line width=0.4pt]
    ($(correct.north west)+(0.25,-0.55)$) -- ($(correct.north east)+(-0.25,-0.55)$);
  \node[align=center,font=\sffamily\scriptsize,text=figInk] at (3.75,0.88)
    {CER\\Challenge-Set\\Keyword Accuracy};

  \node[font=\sffamily\bfseries\scriptsize,text=figCoral]
    at ([yshift=-0.28cm]pause.north) {Pause behavior};
  \draw[draw=figCoral!25,line width=0.4pt]
    ($(pause.north west)+(0.25,-0.55)$) -- ($(pause.north east)+(-0.25,-0.55)$);
  \node[align=center,font=\sffamily\scriptsize,text=figInk] at (7.30,0.88)
    {Pause precision\\PPER\\Intra-word pause rate};

  \node[font=\sffamily\bfseries\scriptsize,text=figTeal]
    at ([yshift=-0.28cm]voice.north) {Voice identity};
  \draw[draw=figTeal!25,line width=0.4pt]
    ($(voice.north west)+(0.25,-0.55)$) -- ($(voice.north east)+(-0.25,-0.55)$);
  \node[align=center,font=\sffamily\scriptsize,text=figInk] at (10.85,0.88)
    {Speaker\\similarity};

  \node[font=\sffamily\bfseries\scriptsize,text=figAmber]
    at ([yshift=-0.28cm]timing.north) {Timing};
  \draw[draw=figAmber!25,line width=0.4pt]
    ($(timing.north west)+(0.25,-0.55)$) -- ($(timing.north east)+(-0.25,-0.55)$);
  \node[align=center,font=\sffamily\scriptsize,text=figInk] at (14.40,0.88)
    {Speaking\\rate};
\end{tikzpicture}}
  \caption{Complementary evaluation for Thai TTS. Synthesized speech is
  scored for sentence correctness, difficult-expression pronunciation, pause
  placement, voice identity, and speaking rate, exposing failures that any single
  metric can miss.}
  \label{fig:evaluation-overview}
\end{figure}
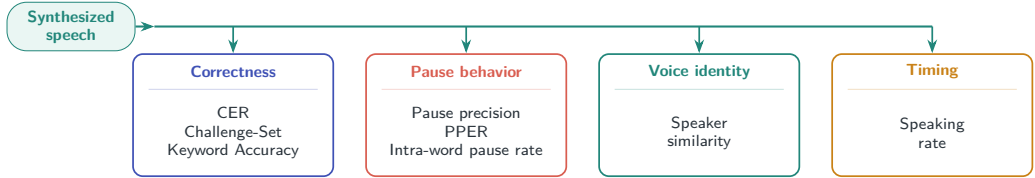

\subsection{Synthetic Speech Corpus Construction}
\label{sec:corpus}

We construct the student corpus in three stages: (1) sourcing and segmenting Thai
text, (2) verbalizing text forms that the multilingual teacher does not reliably
interpret, and (3) sampling and quality-filtering candidate waveforms. The resulting
text--audio pairs are used to train the student. Section~\ref{sec:recipe-ablation}
evaluates these corpus-construction choices.

\subsubsection{Text Sourcing}

Raw text comes from two sources: (1) WangchanThaiInstruct~\citep{limkonchotiwat-etal-2025-wangchanthaiinstruct},
which provides broad Thai coverage, and (2) an LLM-based keyword-synthesis pipeline,
which targets difficult expressions. Before teacher inference, we divide the text into
sentence-level chunks. For the released model we add English text from
LibriTTS~\citep{zen2019libritts}.

\subsubsection{Speech Creation}

We create synthetic speech in two stages. First, we use the OmniVoice Voice Design
mode to generate seed references from 12 speaker specifications. Second, we use the OmniVoice cloning mode
to render corpus texts with these frozen references, producing multiple utterances per
voice while preserving the corresponding identity.

Before cloning, we verbalize text forms that the multilingual teacher does not
reliably interpret. In preliminary experiments, digits were sometimes spoken in
Chinese, while English spans were rendered with an accent that differed from the
intended Thai-accented English pronunciation. The LLM verbalizer therefore rewrites
ambiguous digits and embedded English spans into pronunciation-oriented Thai or
Tinglish text. OmniVoice then synthesizes this prepared text using the frozen seed
reference, and the resulting candidates are passed to the filtering stage.

\subsubsection{Filtering and Rejection Sampling}

To ensure that the synthesized speech is of sufficiently high quality for student-model training, we evaluate each teacher-generated candidate using a quality filter that covers pronunciation, pause placement, speaking rate, and duration.

\paragraph{Content correctness.}
We transcribe each candidate with the CTC-based Thai ASR model
\texttt{airesearch/wav2vec2-large-xlsr-53-th}\footnote{\url{https://huggingface.co/airesearch/wav2vec2-large-xlsr-53-th}}~\citep{airesearch2023wav2vec2thai}
and compare each \emph{hard token} with the target in phoneme space. A candidate is
rejected if any hard token fails an exact phoneme match, including lexical tone. We use
a CTC verifier because Whisper-style\footnote{\url{https://huggingface.co/openai/whisper-large-v3}}
autoregressive ASR models~\citep{radford2022whisper} can recover the intended word from
sentential context despite an incorrect acoustic realization, concealing pronunciation
errors during filtering.

We define hard tokens using deterministic rules. A token is hard if it is
out-of-vocabulary or a TLTK dictionary headword~\citep{aroonmanakun2024tltk} that is
rare in the Thai National Corpus~\citep{aroonmanakun2007tnc,phatthiyaphaibun2023pythainlp}.

\paragraph{Prosody and duration correctness.}
An utterance can pass the content filter while remaining unsuitable for training. We
reject candidates according to three criteria: (1) pause placement that violates the
allowed positions in Section~\ref{sec:pause-eval}, (2) utterance-level speaking rate
outside the speaker-specific range (approximately $\pm10\%$), and (3) hard-token
durations that are abnormally compressed relative to the teacher population.

Pause-placement failures are removed from the candidate pool. We re-render the
remaining failures up to four times and retain the best take if none passes,
preserving text coverage. Table~\ref{tab:filter-rejects} reports the final rejection rate of the filtering process.

\begin{table}[h!]
  \centering
  \small
  \caption{Quality filtering rejects 23.0\% of teacher candidates; pause-placement
  (11.3\%) and pronunciation (10.4\%) failures dominate, while speaking-rate and
  duration checks remove few candidates.}
  \label{tab:filter-rejects}
  \begin{tabular}{lr}
    \toprule
    Quality check & Rejected (\%) \\
    \midrule
    Pause placement & 11.3 \\
    Pronunciation & 10.4 \\
    Speaking rate & 3.0 \\
    Duration & 0.1 \\
    \midrule
    Any rejection & 23.0 \\
    \bottomrule
  \end{tabular}
\end{table}

\subsection{Student Model}
\label{sec:student}

We use the 82M-parameter Kokoro/StyleTTS2 backbone because its compact fixed-voice architecture
matches our deployment objective while retaining strong synthesis
quality~\citep{hexgrad2025kokoro, li2023styletts2}. The student converts phoneme sequences into speech
for a fixed set of voices and does not require reference audio at inference time.
Our adaptation adds a script-routed Thai--English phoneme frontend and trains the model on the
quality-controlled synthetic corpus described in Section~\ref{sec:corpus}.

\subsubsection{Phoneme Frontend}
\label{sec:phoneme-frontend}

Kokoro does not operate directly on raw text. Instead, language-specific frontends
map text to a shared IPA-based phoneme vocabulary; for example, Kokoro uses Misaki
for English and separate frontends for Chinese and Japanese~\citep{hexgrad2025kokoro}.
Because the original frontend does not support Thai, we integrate the Thai Language
Toolkit (TLTK)~\citep{aroonmanakun2024tltk}\footnote{\url{https://pypi.org/project/tltk/}}
as the Thai grapheme-to-phoneme component. Most Thai segmental phones already map to
symbols in Kokoro's multilingual vocabulary. Four of the five Thai lexical tones can
likewise reuse existing contour tokens; only the Thai low tone (เสียงต่ำ) requires an
additional vocabulary entry and a learned embedding.

We compare two treatments of embedded English.

\paragraph{Monolingual frontend.}
Latin spans are converted to pronunciation-oriented Thai script, and TLTK phonemizes
the entire input as Thai.

\paragraph{Bilingual frontend.}
Latin spans remain unchanged; Thai and English spans are routed through TLTK and
Misaki, respectively, and combined in Kokoro's shared phoneme vocabulary. Preserving
English phoneme representations supports transfer to English words, while digits and
symbols remain verbalized in Thai. See Section~\ref{sec:recipe-ablation} for the
corresponding ablation.

\subsubsection{Training}

Unless otherwise stated, we train all models for eight epochs using AdamW~\citep{loshchilov2017adamw}. We use a
learning rate of $1\times10^{-4}$ for the main model and $1\times10^{-5}$ for
PL-BERT\footnote{\url{https://huggingface.co/hexgrad/Kokoro-82M}}~\citep{li2023plbert}. Each optimization step uses an effective batch size of eight.

We initialize PL-BERT, the BERT projection, prosody predictor, text encoder, and
decoder from the released Kokoro checkpoint~\citep{hexgrad2025kokoro}. Since the Kokoro
checkpoint does not include the style encoders required for training, we initialize the
style encoder and predictor encoder from the StyleTTS2-LibriTTS\footnote{\url{https://huggingface.co/yl4579/StyleTTS2-LibriTTS}}
checkpoint~\citep{li2023styletts2}. We also use the pretrained
StyleTTS2 ASR aligner and JDC pitch extractor as training-only supervision modules.
The multi-period and multi-resolution spectrogram discriminators are initialized from
scratch.

For the pretraining ablation, a second student follows the from-scratch initialization protocol of
StyleTTS2~\citep{li2023styletts2}: the modules that StyleTTS2 loads pretrained in every
configuration, including its own from-scratch training -- PL-BERT, the ASR aligner and
the JDC pitch extractor -- remain pretrained, while the BERT projection, prosody
predictor, text encoder, decoder and both style encoders are randomly initialized.

\section{Beyond CER: Evaluating Fixed-Voice Thai TTS}
\label{sec:evaluation}

Standard TTS evaluation commonly reports WER, MOS, and speaker similarity~\citep{hu2026qwen3tts,zhang2025minimaxspeech}. For Thai, WER depends on word segmentation because whitespace does not consistently mark word boundaries. CER avoids this dependency but can obscure critical errors, such as mispronounced names or code-switched expressions. MOS captures overall perceptual quality but requires Thai-specific evaluation infrastructure and provides limited diagnostic insight, while speaker similarity measures voice identity rather than pronunciation or phrasing. We therefore evaluate Thai TTS along four complementary dimensions: (1)~correctness using CER and Challenge-Set Keyword Accuracy, (2)~prosodic phrasing using Prosody Pause Accuracy, (3)~speaker similarity, and (4)~speaking rate, as summarized in Figure~\ref{fig:evaluation-overview}.

\subsection{Pronunciation Correctness}
\label{sec:challenge-eval}

We evaluate content correctness using CER and Challenge-Set Keyword Accuracy. CER measures
sentence-level intelligibility, while Challenge-Set Keyword Accuracy isolates errors on
important local expressions.

$\sbullet$ \textbf{CER.}
We transcribe each utterance using Typhoon Whisper Large V3\footnote{\url{https://huggingface.co/typhoon-ai/typhoon-whisper-large-v3}}~\citep{sirichotedumrong2026typhoonasr} and compute CER after text
normalization and whitespace removal. We report the mean CER, in percent, over a dedicated 500-utterance set
(Section~\ref{sec:eval-data}) containing only Thai-language examples and disjoint
from the Challenge Set. We cap per-utterance CER at 100\% before averaging to
limit the effect of ASR hallucinations, which can otherwise produce arbitrarily
large CER values.

$\sbullet$ \textbf{Challenge-Set Keyword Accuracy.}
We construct a held-out benchmark of 1,531 test sentences across five categories,
as shown in Table~\ref{tab:challenge-set}. Each sentence contains one target expression. An
item is correct when the normalized ASR transcript contains its expected form or an
authorized alternate, with or without spaces. We use exact matching and report both
overall and per-category accuracy.

\begin{table}[h!]
  \centering
  \caption{The 1{,}531-item Challenge Set targets local pronunciation failures that
  sentence-level CER can hide. Each sentence contains one expression from five
  categories and is correct only when the ASR transcript contains an accepted form.}
  \label{tab:challenge-set}
  \begin{tabular}{lr}
    \toprule
    Category & Items \\
    \midrule
    Thai--English code-switching & 391 \\
    Names & 310 \\
    Rare words & 410 \\
    Informal spelling & 210 \\
    Long sentences & 210 \\
    \midrule
    Total & 1,531 \\
    \bottomrule
  \end{tabular}
\end{table}

\begin{figure}[h!]
  \centering
  \resizebox{\textwidth}{!}{\begin{tikzpicture}[
  x=1cm,
  y=1cm,
  font=\sffamily\fontsize{5.4}{6.2}\selectfont,
  >={Stealth[length=2.2mm,width=1.6mm]},
  flow/.style={->,draw=figInk,line width=0.8pt},
  panel/.style={rounded corners=2mm},
  source/.style={draw=figBlue,fill=white,rounded corners=1.4mm,
    minimum width=1.95cm,minimum height=0.58cm,align=center,
    font=\sffamily\bfseries\fontsize{5.2}{6.0}\selectfont,text=figBlue},
  process/.style={draw=figAmber,fill=white,rounded corners=1.5mm,
    minimum width=3.55cm,minimum height=0.68cm,align=center,
    font=\sffamily\bfseries\fontsize{5.2}{6.0}\selectfont,text=figInk},
  input/.style={draw=figMuted!42,fill=white,rounded corners=1.5mm,
    minimum width=2.18cm,minimum height=1.02cm,align=center,
    font=\sffamily\fontsize{5.2}{6.0}\selectfont,text=figInk},
  outcome/.style={rounded corners=1.4mm,minimum width=1.78cm,
    minimum height=0.58cm,align=center,
    font=\sffamily\bfseries\fontsize{5.2}{6.0}\selectfont}
]
  \path[use as bounding box] (0,0.08) rectangle (16,5.00);
  \fill[white] (0,0.08) rectangle (16,5.00);

  \fill[figBluePale,panel] (0.20,0.18) rectangle (5.15,4.82);
  \fill[figAmberPale,panel] (5.35,0.18) rectangle (10.20,4.82);
  \fill[figTealPale,panel] (10.40,0.18) rectangle (15.80,4.82);

  \node[anchor=west,font=\sffamily\bfseries\fontsize{6.2}{7.0}\selectfont,text=figBlue]
    at (0.48,4.49) {CONSTRUCT ITEMS};
  \node[anchor=west,font=\sffamily\bfseries\fontsize{6.2}{7.0}\selectfont,text=figAmber]
    at (5.65,4.49) {SYNTHESIZE + ASR};
  \node[anchor=west,font=\sffamily\bfseries\fontsize{6.2}{7.0}\selectfont,text=figTeal]
    at (10.70,4.49) {EXACT-MATCH TARGET};

  % -----------------------------------------------------------------------
  % 1. Benchmark-item construction
  % -----------------------------------------------------------------------
  \node[source] (curated) at (1.35,3.55) {curated terms};
  \node[source] (mined) at (1.35,2.73) {corpus mining};
  \node[source] (sentences) at (1.35,1.91) {sentence sources};

  \node[draw=figBlue,fill=white,rounded corners=1.7mm,
    minimum width=2.45cm,minimum height=2.85cm,align=center]
    (item) at (3.76,2.73) {};
  \node[font=\sffamily\bfseries\fontsize{5.2}{6.0}\selectfont,text=figBlue]
    at (3.76,3.64) {BENCHMARK ITEM};
  \draw[draw=figBlue!22,line width=0.45pt] (2.76,3.37) -- (4.76,3.37);
  \node[font=\sffamily\fontsize{5.2}{6.0}\selectfont,text=figInk] at (3.76,3.08)
    {test sentence};
  \node[font=\sffamily\fontsize{5.2}{6.0}\selectfont,text=figInk] at (3.76,2.66)
    {target expression};
  \node[font=\sffamily\fontsize{5.2}{6.0}\selectfont,text=figInk] at (3.76,2.24)
    {accepted form(s)};
  \node[draw=figBlue,fill=figBluePale,rounded corners=1.1mm,
    minimum width=1.78cm,minimum height=0.46cm,
    font=\sffamily\bfseries\fontsize{5.2}{6.0}\selectfont,text=figBlue]
    at (3.76,1.58) {1,531 items};

  \draw[flow,draw=figBlue] (curated.east) -- (item.west |- curated.east);
  \draw[flow,draw=figBlue] (mined.east) -- (item.west);
  \draw[flow,draw=figBlue] (sentences.east) -- (item.west |- sentences.east);

  % -----------------------------------------------------------------------
  % 2. System rendering and scoring-ASR transcription
  % -----------------------------------------------------------------------
  \node[process] (render) at (7.78,3.54) {TTS renders the test sentence};
  \node[process] (asr) at (7.78,2.45) {Typhoon Whisper Large V3};
  \node[process] (transcript) at (7.78,1.36) {ASR transcript};

  \draw[flow] (item.east) -- (5.58,2.73) -- (5.58,3.54) -- (render.west);
  \draw[flow,draw=figAmber] (render.south) -- (asr.north);
  \node[font=\sffamily\fontsize{5.0}{5.8}\selectfont,text=figMuted,fill=figAmberPale,inner sep=1pt]
    at (8.50,2.99) {speech};
  \draw[flow,draw=figAmber] (asr.south) -- (transcript.north);

  % -----------------------------------------------------------------------
  % 3. Normalize, exact-match, and record the item verdict
  % -----------------------------------------------------------------------
  \node[input,draw=figBlue] (normalized) at (11.68,3.37)
    {\textbf{\textcolor{figBlue}{normalized transcript}}\\remove spaces};
  \node[input,draw=figTeal] (accepted) at (14.47,3.37)
    {\textbf{\textcolor{figTeal}{accepted forms}}\\expected + alternates};

  \node[draw=figInk,fill=white,rounded corners=1.5mm,
    minimum width=2.80cm,minimum height=0.76cm,align=center,
    font=\sffamily\bfseries\fontsize{5.2}{6.0}\selectfont,text=figInk]
    (match) at (13.08,2.18) {exact substring match?};

  \draw[flow,draw=figBlue] (transcript.east) -- (10.28,1.36)
    -- (10.28,3.37) -- (normalized.west);
  \draw[flow,draw=figBlue] (normalized.south) -- (match.north west);
  \draw[flow,draw=figTeal] (accepted.south) -- (match.north east);

  \node[outcome,draw=figTeal,fill=white,text=figTeal]
    (correct) at (11.86,1.06) {correct item};
  \node[outcome,draw=figCoral,fill=white,text=figCoral]
    (incorrect) at (14.30,1.06) {incorrect item};
  \draw[flow,draw=figTeal] (match.south) -- (13.08,1.61) -- (correct.north);
  \draw[flow,draw=figCoral] (match.south) -- (13.08,1.61) -- (incorrect.north);
  \node[font=\sffamily\fontsize{5.0}{5.8}\selectfont,text=figTeal,fill=figTealPale,inner sep=1pt]
    at (12.18,1.61) {yes};
  \node[font=\sffamily\fontsize{5.0}{5.8}\selectfont,text=figCoral,fill=figTealPale,inner sep=1pt]
    at (13.96,1.61) {no};

  \node[font=\sffamily\bfseries\fontsize{5.2}{6.0}\selectfont,text=figInk]
    at (13.08,0.48) {Keyword Accuracy = correct items / 1,531};
\end{tikzpicture}}
  \caption{Construction and checking of Challenge-Set Keyword Accuracy. Each
  benchmark item pairs a test sentence with one target expression and its accepted
  forms. A system renders the sentence, Typhoon Whisper Large V3 transcribes the
  audio, and the normalized transcript is checked by exact substring matching
  against the expected form or an authorized alternate. We report the fraction of
  correct items overall and by category.}
  \label{fig:keyword-accuracy}
\end{figure}
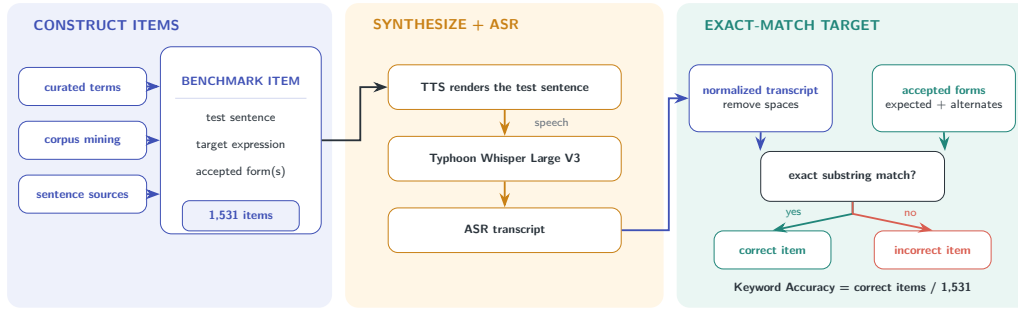

The construction pipeline is described in Appendix~\ref{app:challenge-set} and
summarized with the checking protocol in Figure~\ref{fig:keyword-accuracy}.

\subsubsection{Evaluation Dataset}
\label{sec:eval-data}

We construct the Challenge Set from curated Thai--English terms, sentences from the WangchanThaiInstruct test set~\citep{limkonchotiwat-etal-2025-wangchanthaiinstruct}, VISTEC-TP-TH-2021 annotations~\citep{limkonchotiwat-etal-2021-handling}, PyThaiNLP place names, and a Thai names corpus. Long-sentence items reuse target expressions from the other categories.

The CER set contains 500 transcripts: 250 from the WangchanThaiInstruct test split,
covering retail, finance, medical, and legal domains, and 250 from the Thai validated
test split of Common Voice 17.0~\citep{ardila-etal-2020-common},
covering read-speech prompts. We use only the transcripts and synthesize all evaluation
audio.

\subsection{Prosody Pause Accuracy}
\label{sec:pause-eval}

Text correctness does not imply natural phrasing: an utterance may have an exact
transcript yet pause within a word or at an implausible juncture. This distinction is
especially important in Thai, where whitespace does not reliably mark word boundaries.
We therefore evaluate whether each realized pause occurs at a linguistically
acceptable position, separately from CER and Challenge-Set Keyword Accuracy.

\subsubsection{Evaluation Dataset}
\label{sec:pause-dataset}

Pause placement is evaluated on the 210 long sentences of the Challenge Set,
the subset for which annotated pause masks exist.

\begin{figure}[h!]
  \centering
  \resizebox{\textwidth}{!}{\begin{tikzpicture}[
  x=1cm,
  y=1cm,
  font=\sffamily,
  >={Stealth[length=2.2mm,width=1.6mm]},
  flow/.style={->,draw=figInk,line width=0.8pt},
  panel/.style={rounded corners=2mm},
  example/.style={draw=figMuted!38,fill=white,rounded corners=1.5mm,
    minimum width=5.25cm,minimum height=1.40cm,inner sep=4pt},
  source/.style={draw=figMuted!42,fill=white,rounded corners=1.5mm,
    minimum width=2.55cm,minimum height=0.60cm,align=center,
    font=\sffamily\bfseries\scriptsize},
  token/.style={draw=figMuted!32,fill=figMuted!8,rounded corners=0.7mm,
    minimum width=0.40cm,minimum height=0.38cm,inner sep=0pt}
]
  \path[use as bounding box] (0,0.62) rectangle (16,4.80);
  \fill[white] (0,0) rectangle (16,4.80);

  % Two large regions keep the comparison and construction visually separate.
  \fill[figBluePale,panel] (0.20,0.78) rectangle (6.18,4.58);
  \fill[figTealPale,panel] (6.42,0.78) rectangle (15.80,4.58);

  \node[anchor=west,font=\sffamily\bfseries\small,text=figBlue]
    at (0.48,4.25) {ORTHOGRAPHIC PAUSE CUES};
  \node[anchor=west,font=\sffamily\bfseries\small,text=figTeal]
    at (6.72,4.25) {THAI PAUSE EVALUATION SET};

  % English example: the punctuation itself supplies the visible cue.
  \node[example] (english) at (3.19,3.28) {};
  \draw[draw=figBlue,line width=2.0pt] (0.62,3.51) -- (0.62,3.88);
  \node[anchor=west,font=\sffamily\bfseries\scriptsize,text=figBlue]
    at (0.76,3.70) {English};
  \node[font=\sffamily\small,text=figInk]
    at (3.19,3.31)
    {Please wait{\color{figBlue}\bfseries ,} then continue.};
  \node[anchor=west,font=\sffamily\scriptsize,text=figBlue]
    at (1.72,2.86) {punctuation $\longrightarrow$ pause cue};

  % Thai example: the annotation can expose a boundary absent from the string.
  \node[example] (thai) at (3.19,1.62) {};
  \draw[draw=figAmber,line width=2.0pt] (0.62,1.85) -- (0.62,2.22);
  \node[anchor=west,font=\sffamily\bfseries\scriptsize,text=figAmber]
    at (0.76,2.04) {Thai};
  \node[anchor=east,font=\thaifont\footnotesize,text=figInk]
    at (3.15,1.64) {แม้ฝนจะตกหนัก};
  \node[anchor=west,font=\thaifont\footnotesize,text=figInk]
    at (3.15,1.64) {เราก็ยังเดินทางต่อ};
  \draw[draw=figTeal,line width=1.0pt,dashed]
    (3.15,1.42) -- (3.15,1.82);
  \node[anchor=west,font=\sffamily\scriptsize,text=figTeal]
    at (1.72,1.12) {acceptable pause; no written mark};

  % Three explicit Thai annotation sources.
  \node[source,draw=figBlue,text=figBlue] (space) at (7.97,3.40)
    {source space};
  \node[source,draw=figAmber,text=figAmber] (punct) at (7.97,2.57)
    {punctuation};
  \node[source,draw=figTeal,text=figTeal] (context) at (7.97,1.74)
    {contextual mask};

  \node[draw=figInk,fill=white,rounded corners=1.4mm,
    minimum width=1.28cm,minimum height=0.72cm,align=center,
    font=\sffamily\bfseries\scriptsize,text=figInk] (union)
    at (10.35,2.57) {UNION};

  \draw[flow,draw=figBlue] (space.east) -- (9.32,3.40) -- (union.north west);
  \draw[flow,draw=figAmber] (punct.east) -- (union.west);
  \draw[flow,draw=figTeal] (context.east) -- (9.32,1.74) -- (union.south west);

  % Output annotation: abstract word tiles with allowed boundary marks.
  \node[draw=figTeal,fill=white,rounded corners=1.7mm,
    minimum width=4.30cm,minimum height=3.05cm,align=center]
    (mask) at (13.44,2.57) {};
  \draw[flow,draw=figInk] (union.east) -- (mask.west);

  \node[font=\sffamily\bfseries\scriptsize,text=figTeal]
    at (13.44,3.55) {ALLOWED-POSITION MASK};

  \foreach \x in {11.78,12.32,12.86,13.40,13.94,14.48,15.02}
    \node[token] at (\x,2.89) {};
  \draw[draw=figBlue,line width=1.4pt] (12.59,2.59) -- (12.59,3.19);
  \draw[draw=figAmber,line width=1.4pt] (13.67,2.59) -- (13.67,3.19);
  \draw[draw=figTeal,line width=1.4pt,dashed] (14.75,2.59) -- (14.75,3.19);

  \node[font=\sffamily\scriptsize,text=figInk,align=center]
    at (13.44,2.22) {210 long Thai sentences\\+ boundary masks};
  \node[draw=figTeal,fill=figTealPale,rounded corners=1.2mm,
    minimum width=3.18cm,minimum height=0.54cm,
    font=\sffamily\bfseries\scriptsize,text=figTeal]
    at (13.44,1.47) {Prosody Pause evaluation set};

\end{tikzpicture}}
  \caption{Construction of the Prosody Pause evaluation set. English punctuation
  often provides a direct pause cue (left, top), whereas a linguistically acceptable
  Thai phrase boundary may be unmarked (left, bottom). For each of the 210 long Thai
  sentences, we therefore take the union of author-written spaces, punctuation, and
  contextual boundaries from the text-only mask. The resulting annotation specifies
  allowed, rather than required, pause positions.}
  \label{fig:pause-mask-construction}
\end{figure}
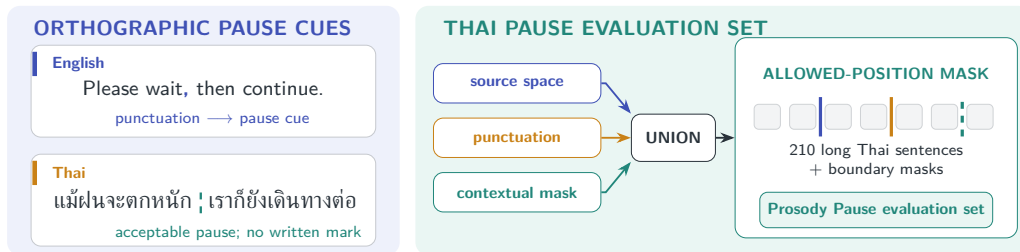

The set of acceptable positions has two parts. The first is derived only from the
input text: a source space or a punctuation mark, both of which the author actually
wrote in the text. The second is a per-sentence mask. We prompt the text-only
\texttt{gemini-3.1-pro-preview}\footnote{\url{https://ai.google.dev/gemini-api/docs/models/gemini-3.1-pro-preview}} model to mark every position at which a
pause would be acceptable similar to \citep{geng-etal-2025-scaling}. The allowed set is the union of the two parts, as
illustrated in Figure~\ref{fig:pause-mask-construction}.

\subsubsection{Scoring Method}

Pause placement is a set-valued prediction problem: several boundaries may permit a
pause, but fluent speech need not realize any particular one. The reference therefore
specifies \emph{allowed}, rather than required, positions. Recall is not meaningful
under this contract. Instead, we report (1) pause precision, the fraction of detected
pauses at allowed positions; (2) pause-placement error rate (PPER), the fraction of
clips containing at least one misplaced pause; and (3) intra-word pause rate, the
fraction containing the most severe error class. Because a system can improve PPER by
pausing less often, we report pauses per clip alongside these measures and use pause
precision as the primary placement measure.

\begin{figure}[h!]
  \centering
  \includegraphics[width=\linewidth]{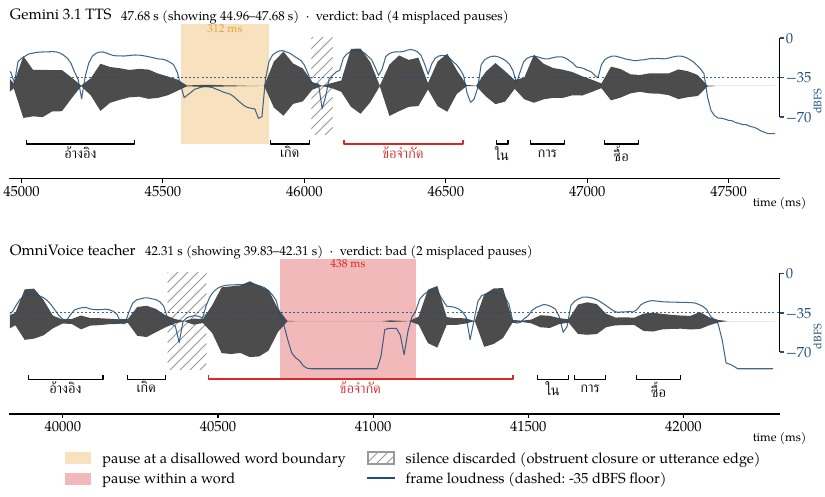}
  \caption{ASR transcripts can conceal pause errors. On
  \emph{ดัชนีอ้างอิงเกิดข้อจำกัดในการซื้อ} (``index tracking creates a constraint on
  buying''), OmniVoice (bottom) swallows most of \emph{ข้อจำกัด} (``constraint'') in a
  438~ms intra-word pause, while Gemini 3.1 (top) renders the word continuously. Both
  systems transcribe \emph{ข้อจำกัด} correctly, so the split leaves the word's
  transcript unchanged. Word positions use CTC character timestamps.}
  \label{fig:intra-word-pause}
\end{figure}

The scorer detects internal silent regions, excludes likely consonant closures and
silences the alignment cannot account for, and maps each remaining pause to the input
text using TLTK segmentation and forced alignments from the external Thai CTC ASR model
\texttt{airesearch/wav2vec2-large-xlsr-53-th}~\citep{airesearch2023wav2vec2thai}.
It then classifies the pause as allowed, at an implausible word boundary, or within a
word. 
We also evaluated another ASR aligner during validation. Typhoon ASR CTC (Whisper)\footnote{\url{https://huggingface.co/typhoon-ai/typhoon-whisper-large-v3-ctc}} performs strongly, achieving the highest clip-verdict agreement (90.7\%) and a per-speaker pause-precision correlation of 0.975 with the duration-predictor reference. See Appendix~\ref{app:pause-params} for details on the metrics and validation procedure.

Figure~\ref{fig:intra-word-pause} shows an example: the same word is rendered
continuously by one reference system and split by an intra-word pause in the other,
while both transcribe it correctly.

\subsection{Speaker Similarity and Speaking Rate}

\subsubsection{Speaker Similarity}
\label{sec:speaker-eval}

We evaluate speaker similarity on the 210 long sentences from the
Challenge Set. For each voice, we extract speaker embeddings using
\texttt{speechbrain/spkrec-ecapa-voxceleb}\footnote{\url{https://huggingface.co/speechbrain/spkrec-ecapa-voxceleb}}~\citep{desplanques2020ecapa}
and compute cosine similarity to the target voice centroid, constructed from up
to 40 teacher-generated training utterances for each speaker. We report mean cosine similarity on $[0,1]$, where higher is better and 1
indicates identical speaker embeddings.

\subsubsection{Speaking Rate}
\label{sec:rate-eval}

We measure speaking rate in tokens per voiced second on the 210 long sentences from the Challenge Set.
For each voice, we report how much faster or slower the student
speaks than its own teacher reference, as a percentage of the teacher rate.

\subsection{Why CER Is Not Enough}

This experiment examines whether CER alone captures targeted pronunciation and
pause-placement errors. We compare Gemini 3.1 Flash
TTS\footnote{\url{https://ai.google.dev/gemini-api/docs/models/gemini-3.1-flash-tts-preview}} with the OmniVoice teacher.

\begin{table}[h!]
  \centering
  \caption{Gemini 3.1 improves both mean CER (3.3\% vs.\ 4.6\%) and Challenge-Set
  Keyword Accuracy (79.8\% vs.\ 72.8\%) over OmniVoice. Keyword scoring directly
  exposes errors on difficult local expressions that sentence-level CER averages over.}
  \label{tab:cer-keyword}
  \begin{tabular}{lcc}
    \toprule
    System & CER mean & Keyword Acc. \\
    \midrule
    Gemini 3.1 Flash TTS & 3.3\% & 79.8\% \\
    OmniVoice teacher & 4.6\% & 72.8\% \\
    \bottomrule
  \end{tabular}
\end{table}

As shown in Table~\ref{tab:cer-keyword}, Gemini 3.1 Flash TTS outperforms
OmniVoice on both measures. OmniVoice has a higher CER (4.6\% vs.\ 3.3\%) and
a 7.0-point lower Challenge-Set Keyword Accuracy (72.8\% vs.\ 79.8\%).
The difference in keyword accuracy shows that sentence-level CER does not fully
characterize errors on challenging expressions.

\begin{table}[h!]
  \centering
  \caption{Pause placement on 210 long sentences. Gemini 3.1 attains higher precision
  and fewer clips with misplaced (PPER) or intra-word pauses than OmniVoice, despite
  producing more than twice as many pauses per clip.}
  \label{tab:pause-comparison}
  \begin{tabular}{lcccc}
    \toprule
    System & Pause Prec. & PPER & Intra-word & Pauses/clip \\
    \midrule
    Gemini 3.1 Flash TTS& \textbf{96.4\%} & \textbf{13.8\%} & \textbf{1.9\%} & 4.44 \\
    OmniVoice teacher & 89.9\% & 17.6\% & 5.2\% & 1.79 \\
    \bottomrule
  \end{tabular}
\end{table}

As shown in Table~\ref{tab:pause-comparison}, the pause results reveal a similar distinction. Gemini 3.1 Flash TTS achieves
higher pause precision, lower PPER, and a lower intra-word pause rate than
OmniVoice. Importantly, it does so while producing more than twice as many pauses
per clip (4.44 vs.\ 1.79). This difference in pause frequency matters when
interpreting PPER: because PPER records whether a clip contains at least one
misplaced pause, a system that pauses less often has fewer opportunities to
incur such an error. Despite pausing substantially less often, OmniVoice still
has the higher PPER (17.6\% vs.\ 13.8\%).

Together, these results illustrate why CER alone is insufficient for evaluating
Thai TTS. Challenge-Set Keyword Accuracy identifies errors on difficult target
expressions, while Prosody Pause Accuracy captures errors in phrasing that may
leave the transcript unchanged.

\section{Building the Final Model}
\label{sec:results}

\subsection{Recipe Ablation}
\label{sec:recipe-ablation}

\begin{table*}[h!]
  \centering
  \fontsize{8}{9.6}\selectfont
  \setlength{\tabcolsep}{1.5pt}
  \caption{Cumulative recipe ablation; each row adds the named component to the
  preceding configuration. Resampling rejected texts preserves hard-example coverage
  after filtering, yielding the best CER and pause metrics with near-best keyword
  accuracy and unchanged speaker similarity. Bold marks strict metric optima. We refer
  to the last configuration as the Thai-only model.}
  \label{tab:recipe-ablation}
  \begin{tabular}{l@{\hspace{6pt}}rrrrrrrr}
    \toprule
    Corpus construction & Hours & Keyword & CER & Pause & PPER &
    Intra-word & Speaker & Rate \\
    & & $\uparrow$ & mean $\downarrow$ & Prec. $\uparrow$ &
    $\downarrow$ & $\downarrow$ & sim.\ $\uparrow$ & dev.\ \% $\downarrow$ \\
    \midrule
    Unfiltered monolingual baseline
      & 17.72 & 67.5\% & 4.0\% & 81.3\% & 16.2\% & 4.3\% & 0.882 & 12.4 \\
    $+$ Pause filtering
      & 17.63 & \textbf{69.6\%} & 3.5\% & 88.4\% & 14.3\% & 5.7\% & 0.875 & 8.2 \\
    $+$ Bilingual frontend
      & 17.63 & 68.7\% & 4.1\% & 88.2\% & 14.8\% & 5.7\% & 0.878 & \textbf{8.0} \\
    $+$ Quality-filtered scale-up
      & 40.58 & 67.2\% & 3.9\% & 85.4\% & 11.9\% & 3.3\% & 0.881 & 12.2 \\
    $+$ Resample rejected candidates
      & 41.81 & 69.2\% & \textbf{3.4\%} & \textbf{92.8\%} & \textbf{6.2\%} & \textbf{1.4\%} & 0.882 & 9.5 \\
    \bottomrule
  \end{tabular}
\end{table*}

This experiment evaluates corpus-construction decisions across correctness, phrasing,
voice identity, and speaking rate. Table~\ref{tab:recipe-ablation} builds the recipe
cumulatively, with each row adding one component to the preceding configuration.
All experiments use a matched sampling budget: we oversample outputs from the teacher
model and then retain valid examples through sampling or filtering according to the
criteria described in Table~\ref{tab:filter-rejects}.

At approximately 17.6 hours of training data, we first evaluate the effect of pause
filtering. Relative to the unfiltered monolingual baseline, pause filtering increases
Keyword Accuracy from 67.5\% to 69.6\% and reduces mean CER from 4.0\% to 3.5\%.
Pause precision also improves from 81.3\% to 88.4\%, while speaking-rate deviation
decreases from 12.4\% to 8.2\%. These results suggest that removing teacher outputs
with problematic pause placement not only improves phrasing but also produces cleaner
training targets for content correctness.

We next introduce the bilingual frontend while keeping the training corpus fixed at
17.63 hours. This frontend allows Latin spans to remain in their original form rather
than requiring them to be verbalized into Thai, and it reuses pretrained Kokoro embeddings for supported non-Thai symbols. Keyword Accuracy remains close to the
pause-filtered configuration (68.7\% vs.\ 69.6\%), although mean CER increases from
3.5\% to 4.1\%. Pause behavior is essentially unchanged: pause precision moves from
88.4\% to 88.2\%, PPER from 14.3\% to 14.8\%, and the intra-word pause rate is flat at
5.7\%. Thus, at a fixed corpus size, the bilingual frontend preserves phrase-boundary behavior while allowing Latin spans to be read as written.

Expanding the corpus from 17.63 to 40.58 hours under the content filter, in
addition to pause filtering, slightly improves mean CER and reduces PPER and
intra-word errors. However, Keyword Accuracy falls from 68.7\% to
67.2\%, pause precision from 88.2\% to 85.4\%, and speaking-rate deviation increases
from 8.0\% to 12.2\%. The decline
in Keyword Accuracy may result from the repeated filtering of difficult examples,
which reduces their representation in the training corpus.

Resampling rejected candidates then reverses these
trade-offs: relative to scale-up without resampling, it improves Keyword Accuracy by
2.0 points, reduces mean CER by 0.5 points, improves every pause measure, and reduces
speaking-rate deviation while leaving speaker similarity unchanged. The final recipe consequently
achieves the best CER and pause placement among the tested configurations, retains near-best Keyword Accuracy, and preserves the target voice.

In summary, synthetic-corpus construction must balance quality and coverage. A strict
filter removes defective teacher outputs, but simply discarding rejected examples also
removes difficult training texts. Resampling instead searches for an acceptable realization of each rejected text, thereby preserving coverage. Among the tested
configurations, resampling rejected examples produces the broadest overall improvement.
We therefore use the complete resampling recipe to train our final model,
\textbf{Wayu-Paxa-TTS-Edge}.

\subsection{Effect of Pretrained Initialization}

\begin{table}[h!]
  \centering
  \caption{Pretrained initialization with Kokoro and StyleTTS2 modules improves every reported
  correctness and pause metric; the largest change is code-switch accuracy, from
  22.8\% to 65.5\%.}
  \label{tab:pretraining}
  \begin{tabular}{lrr}
    \toprule
    Metric & From scratch & Pretrained \\
    \midrule
    CER mean & 6.6\% & 3.4\% \\
    Keyword Acc. & 51.7\% & 69.2\% \\
    Code-switch Acc. & 22.8\% & 65.5\% \\
    Long-sentence Acc. & 58.1\% & 72.4\% \\
    Name Acc. & 47.1\% & 53.2\% \\
    Pause Prec. & 82.0\% & 92.8\% \\
    PPER & 12.9\% & 6.2\% \\
    Intra-word pause & 3.8\% & 1.4\% \\
    \bottomrule
  \end{tabular}
\end{table}

This experiment isolates the effect of student initialization. The model with pretrained initialization
loads PL-BERT, the BERT projection, prosody predictor, text encoder, and decoder from
Kokoro~\citep{hexgrad2025kokoro}, and the style encoder and predictor encoder from
StyleTTS2. The from-scratch model initializes the BERT projection, prosody predictor,
text encoder, decoder and both style encoders from scratch, and keeps PL-BERT, the ASR
aligner and the JDC pitch extractor pretrained, following StyleTTS2's protocol.

Pretrained initialization improves CER, keyword accuracy, and pause placement. The largest
keyword gain is observed for code-switching ($+43.4$ points), while the gain for
names is substantially smaller ($+6.1$ points). Overall, reusing the pretrained
Kokoro model consistently outperforms training from scratch across all evaluation
metrics.

\subsection{Frontend Changes Improve Performance Without Retraining}

This experiment asks how much Challenge-Set error can be corrected at inference time
without modifying the acoustic model. We evaluate two frontend interventions on the
held-out 1{,}531-item Challenge Set from Section~\ref{sec:eval-data}: LLM-based
verbalization and expanded phoneme handling in the TLTK-based G2P. We implement
verbalization with
\texttt{deepseek-ai/DeepSeek-V4-Flash}\footnote{\url{https://huggingface.co/deepseek-ai/DeepSeek-V4-Flash}}
and apply both interventions cumulatively using fixed epoch-8 weights from the Thai-only
model. Table~\ref{tab:frontend-only} reports the results.

\begin{table}[h!]
  \centering
  \caption{Inference-only LLM verbalization and broader TLTK phoneme handling affect
  12.3\% and 7.3\% of inputs, respectively, but together raise Challenge-Set Keyword
  Accuracy from 68.9\% to 70.0\% without acoustic-model retraining.}
  \label{tab:frontend-only}
  \small
  \setlength{\tabcolsep}{3pt}
  \begin{tabular}{lrrr}
    \toprule
    Change & Inputs changed & \multicolumn{2}{c}{Keyword Acc.} \\
    \cmidrule(lr){3-4}
    & & Before & After \\
    \midrule
    LLM verbalization & 12.3\% & 68.9\% & 69.2\% \\
    \shortstack[l]{TLTK G2P: improve phoneme handling} & 7.3\% & 69.2\% & 70.0\% \\
    \bottomrule
  \end{tabular}
\end{table}

Table~\ref{tab:frontend-only} shows that final Keyword Accuracy can be improved through changes to text normalization and frontend processing alone. However, the gains are not proportional to how often each frontend component modifies the input. LLM-based verbalization changes 12.3\% of examples but improves Keyword Accuracy by only 0.3 percentage points, whereas expanded G2P handling affects 7.3\% yet produces a larger 0.8-point gain. This asymmetry suggests that, among the cases addressed by these interventions, residual errors are more sensitive to pronunciation representation than to broad text verbalization. More generally, recovering 1.1 accuracy points with fixed model weights indicates that a measurable portion of Challenge-Set failures originates upstream of acoustic generation. Keeping normalization and pronunciation handling outside the acoustic model therefore provides a practical way to address long-tail Thai text cases without retraining the TTS model.

\subsection{Final Model Performance}
This experiment evaluates whether the compact student preserves its Thai synthesis
behavior after adding English while remaining competitive with larger TTS systems. We
compare the released \textbf{Wayu-Paxa-TTS-Edge} checkpoint with its OmniVoice teacher
and Gemini 3.1 Flash TTS. Starting from the Thai recipe in
Section~\ref{sec:recipe-ablation}, we add English utterances synthesized from LibriTTS
transcripts using the same twelve reference speakers and filtering pipeline. We cap
English data at 30\% of the Thai corpus duration, yielding 54.35 hours of bilingual
training data. Table~\ref{tab:final-model} reports the results.

Despite its 82M-parameter backbone, Wayu-Paxa-TTS-Edge preserves most of the teacher's
Thai synthesis behavior. It remains close to OmniVoice in speaker similarity, improves
Thai CER, and produces fewer pause errors, including the lowest PPER and intra-word
pause rate among the compared systems. The main degradation appears on the Challenge
Set, where Keyword Accuracy trails both OmniVoice and Gemini. This result suggests that
compression primarily affects difficult lexical realization rather than speaker
preservation or pause structure.

Adding English introduces a similarly localized trade-off. Relative to the Thai-only
model, English CER decreases from 4.4\% to 1.1\%, while speaker similarity and the
intra-word pause rate remain unchanged. The main costs are modest reductions in
Challenge-Set Keyword Accuracy and pause precision. Overall, the bilingual checkpoint
adds strong English capability without broadly degrading Thai synthesis, supporting its
use as the deployment model.

\begin{table*}[h!]
  \centering
  \fontsize{8}{9.6}\selectfont
  \setlength{\tabcolsep}{2.5pt}
  \caption{Final comparison: the 82M-parameter student nearly matches larger
  systems in CER, surpasses its teacher in pause precision, and attains the lowest PPER
  and intra-word pause rate among the three systems. Keyword accuracy and Thai CER use 1{,}531 and 500 items,
  respectively; English CER uses 500 held-out LibriTTS lines transcribed with Whisper
  Large~V3~\citep{radford2022whisper}; pause and speaker
  metrics use 210 long sentences (Gemini similarity uses Kore references). The Thai-only
  model is the Table~\ref{tab:recipe-ablation} recipe endpoint, shown for reference.
  Pauses per clip contextualizes pause frequency.}
  \label{tab:final-model}
  \begin{tabular}{lrrrrrrrrr}
    \toprule
    System & Params. $\downarrow$ & Keyword $\uparrow$ & \shortstack{CER\\Thai $\downarrow$} &
    \shortstack{CER\\Eng. $\downarrow$} &
    \shortstack{Pause\\prec. $\uparrow$} & PPER $\downarrow$ &
    \shortstack{Intra-\\word $\downarrow$} & \shortstack{Pauses\\per clip} &
    \shortstack{Speaker\\sim. $\uparrow$} \\
    \midrule
    \multicolumn{10}{l}{\emph{Open-weight models}} \\
    \ourrow Wayu-Paxa-TTS-Edge & \textbf{82M} & 68.2\% & 3.7\% & 1.1\% & 91.4\% & 6.7\% &
      \textbf{1.4\%} & 0.89 & 0.882 \\
    \quad \emph{Thai-only model} & \textbf{82M} & 69.2\% & \textbf{3.4\%} & 4.4\% & \textbf{92.8\%} & \textbf{6.2\%} &
      \textbf{1.4\%} & 0.99 & 0.882 \\
    OmniVoice teacher & 600M & \textbf{72.8\%} & 4.6\% & \textbf{0.9\%} & 89.9\% & 17.6\% & 5.2\% & 1.79 & \textbf{0.899} \\
    \midrule
    \multicolumn{10}{l}{\emph{Proprietary model}} \\
    Gemini 3.1 Flash TTS & $\geq$405B\textsuperscript{*} & 79.8\% & 3.3\% & 0.8\% & 96.4\% & 13.8\% & 1.9\% & 4.44 & 0.816 \\
    \bottomrule
  \end{tabular}
  \vspace{2pt}
  {\footnotesize\textsuperscript{*}Literature-derived lower-bound proxy for Gemini
  3.1 Flash TTS~\citep{nikolic2026inferringsize,li2026incompressibleknowledge}.}
\end{table*}

\subsection{Toward Isan Dialect Adaptation from a 15-Second Reference}
\label{sec:isan-adaptation}

This experiment tests whether a short Isan-accented reference can specify a new fixed
voice. From a 15-second crop, OmniVoice renders approximately 1.5 hours of Isan text, and we fine-tune the Thai-only model for four Isan-only
epochs. The text comes from the \texttt{isan\_spelling} field of the Thai Dialect
Isan Speech Corpus training split. Because the
Central-Thai CTC verifier rejects dialect forms themselves, we disable the
pronunciation filter while retaining the prosody and duration filters.

\begin{table}[h!]
  \centering
  \caption{Four-epoch adaptation on approximately 1.5 hours of synthetic Isan speech
  from a 15-second reference improves on teacher Isan CER (5.5\% vs.\ 6.6\%) and nearly
  retains voice identity (0.842 vs.\ 0.854), but degrades Central-Thai CER and keyword
  accuracy. Similarity uses 150 clips; Isan and Central-Thai CER use both 500 items.}
  \label{tab:isan-adaptation}
  \begin{tabular}{lrrr}
    \toprule
    Metric & Teacher & Before & After \\
    \midrule
    Speaker cosine similarity $\uparrow$ & 0.854 & -- & 0.842 \\
    Full-test Isan CER $\downarrow$ & 6.6\% & -- & 5.5\% \\
    Central-Thai CER $\downarrow$ & -- & 3.4\% & 5.1\% \\
    Central-Thai Keyword Accuracy $\uparrow$ & -- & 69.2\% & 65.8\% \\
    \bottomrule
  \end{tabular}
\end{table}

The adapted student's cosine similarity to the reference is 0.842, close to the
teacher's 0.854. On 500-item Isan\footnote{\url{https://huggingface.co/datasets/typhoon-ai/thai-dialect-isan-dataset}} test split, the
dialect-specific ASR\footnote{\url{https://huggingface.co/typhoon-ai/typhoon-isan-asr-whisper}} gives the student a CER of 5.5\%,
compared with 6.6\% for the teacher. On Central Thai, CER increases from 3.4\% to 5.1\%,
while Challenge-Set Keyword Accuracy decreases from 69.2\% to 65.8\%. These results
show that a 15-second reference can specify a fixed voice. While CER measures
intelligibility, it does not
establish Isan naturalness or dialect fidelity; a comprehensive Isan evaluation set is
therefore a natural next step.

\section{Teacher Analysis and the Limits of Distillation}

Having established the final model's performance, we consider a direct opportunity to
improve the student's Challenge-Set Keyword Accuracy
(Section~\ref{sec:eval-data}): capturing more of the teacher's usable distribution
through repeated sampling. We first measure the recoverable gains, then examine where
limited teacher support constrains them.

\subsection{Recovery through Repeated Sampling}

This experiment tests how much repeated teacher sampling can recover. We use oracle
best-of-$K$ selection to measure whether the teacher can produce a correct realization.

\begin{table}[h!]
  \centering
  \caption{Oracle best-of-$K$ teacher sampling on the Challenge Set. Repeated sampling
  raises exact accuracy from 72.8\% at $K=1$ to 87.9\% at $K=118$, revealing 15.1
  percentage points of recoverable headroom relative to one sample.}
  \label{tab:oracle-sampling}
  \begin{tabular}{rr}
    \toprule
    $K$ & Exact Acc. \\
    \midrule
    1 & 72.8\% \\
    6 & 81.4\% \\
    22 & 84.8\% \\
    54 & 86.7\% \\
    86 & 87.3\% \\
    118 & 87.9\% \\
    \bottomrule
  \end{tabular}
\end{table}

Best-of-$K$ selection improves exact accuracy from 72.8\% to 87.9\%, showing that
repeated sampling recovers many correct realizations not obtained from one sample.

\subsection{Training-Corpus Coverage and Sampling Difficulty}

We analyze the OmniVoice training data to contextualize oracle recovery. OmniVoice
reports approximately 10.5k hours of Thai speech, of which 98.2\% comes from
GigaSpeech~2~\citep{yang-etal-2025-gigaspeech,zhu2026omnivoice}. Oracle recovery
reflects this distribution: keyword frequency correlates with sampling difficulty
($\rho=-0.437$).

\begin{table}[h!]
  \centering
  \caption{Teacher-corpus frequency correlates with sampling difficulty across 1{,}060
  Thai only keywords ($\rho=-0.437$). At $K=118$, coverage is 66\% for unseen
  keywords but 98\% for those observed 100--10k times.}
  \label{tab:teacher-frequency}
  \small
  \begin{tabular}{lrrrr}
    \toprule
    Occurrences & Items & $K=6$ & $K=118$ & Unresolved \\
    \midrule
    0 (unseen) & 286 & 48\% & 66\% & 21\% \\
    1--9 & 134 & 55\% & 86\% & 7\% \\
    10--99 & 110 & 82\% & 96\% & 4\% \\
    100--999 & 248 & 94\% & 98\% & 1\% \\
    1k--10k & 244 & 95\% & 98\% & 2\% \\
    10k+ & 38 & 82\% & 92\% & 5\% \\
    \bottomrule
  \end{tabular}
\end{table}

At $K=118$, coverage rises from 66\% for unseen keywords to 98\% for keywords observed
100--10k times.

Rejection sampling therefore expands the usable teacher distribution, while the
teacher's existing support determines how far that expansion can go.

\section{Discussion, Limitations, and Conclusion}
\label{sec:discussion}
Our results show that smaller TTS models can remain competitive with large proprietary models despite using substantially more compact architectures than current zero-shot voice-cloning models. However, some evaluation metrics may favor larger models, and a performance gap with open-source teacher models remains.

From an engineering perspective, pause and keyword accuracy provide useful signals for corpus construction and model evaluation, but remain imperfect due to their reliance on heuristics, including Thai rule-based tokenization and phoneme conversion. Given the context-dependent nature of Thai, combining neural and rule-based approaches may be a promising direction for future work.

We hope this work encourages further research on Thai speech technology and other languages facing similar challenges.

\section*{Acknowledgments}

This work is a collaboration between Wayu Research, Paxa Labs, and Typhoon. It was self-funded by Wayu Research. We thank the Typhoon team for releasing Typhoon-ASR-CTC\footnote{\url{https://huggingface.co/typhoon-ai/typhoon-whisper-large-v3-ctc}}, as well as the global and local AI communities for open-sourcing resources and sharing knowledge that made this work possible.

\section*{Ethics Statement}
Synthetic voice construction can reduce the cost of building speech systems for
resource-constrained languages, but it can also enable unauthorized voice imitation. The
voice references used for training and evaluation should have clear provenance and
appropriate consent or licensing. Released artifacts should document intended uses,
known pronunciation and accent limitations, and safeguards against impersonation.

\bibliography{colm2026_conference}
\bibliographystyle{colm2026_conference}

\appendix

\section{Additional Experimental Details}
\label{app:details}

\subsection{Challenge-Set Construction}
\label{app:challenge-set}

Construction metadata are included with the released
benchmark.\footnote{\url{https://github.com/wayu-research/thai-tts-eval}} For mined Thai
targets, we retain dictionary headwords or true out-of-vocabulary forms that are rare
in the Thai National Corpus~\citep{aroonmanakun2007tnc} and are either exceptionally
rare or orthographically irregular. We treat silent-letter marks, written clusters,
irregular final consonants, and linking-syllable junctures as irregularities, and
exclude compounds of common components unless the compound exhibits one of them. For
mined Latin-script targets, we retain forms that are all-uppercase, mixed-case,
digit-containing, or absent from the English pronunciation lexicon.

After target selection, we remove tokenizer fragments and non-words. We retain a
corpus sentence when one is available and author a sentence otherwise, filtering
corpus sentences by length, Thai-character share, and digit share.
Table~\ref{tab:challenge-construction} summarizes the remaining category-specific
construction details.

\begin{table}[h!]
  \centering
  \caption{Category-specific target selection and stratification details.}
  \label{tab:challenge-construction}
  \small
  \setlength{\tabcolsep}{4pt}
  \begin{tabular}{lr>{\raggedright\arraybackslash}p{9.0cm}}
    \toprule
    Category & Items & Construction details \\
    \midrule
    Code-switching & 391 & Curated SET50/SET100 tickers and Thai brands, together with mined
    Latin tokens; stratified as acronyms, loanwords, brands, and products \\
    Names & 310 & Personal names from a Thai names corpus, PyThaiNLP
    provinces~\citep{phatthiyaphaibun2023pythainlp}, and VISTEC named-entity spans \\
    Rare words & 410 & Dictionary headwords or true out-of-vocabulary forms, stratified by
    Thai National Corpus frequency \\
    Informal spelling & 210 & VISTEC observed/canonical spelling pairs \\
    Long sentences & 210 & Targets reused from other categories, stratified by sentence
    length and target position (early, middle, or final) \\
    \midrule
    Total & 1{,}531 & \\
    \bottomrule
  \end{tabular}
\end{table}

\subsection{Prosody Pause Metric}
\label{app:pause-params}

\paragraph{Scope and residual error.} The scorer is a composite of several components --
Thai word segmentation, CTC forced alignment, and the handling of code-switched and
verbalized spans -- and its accuracy is bounded by all of them. Segmentation decides what
counts as word-internal, and the two tokenizers we combine still agree on some compounds
that a listener would treat as two words; the character aligner contributes timing
jitter, and it has no vocabulary for digits or Latin spans, whose spoken Thai form it
therefore cannot place. The guards described below remove the failure modes we
identified, but the detector is not exact. Its rates are most reliable as comparisons
between systems measured with the same instrument, and reducing these component
dependencies is left to future work.

Table~\ref{tab:pause-hyperparams} lists the fixed implementation constants omitted
from Section~\ref{sec:pause-eval}.

\begin{table}[h!]
  \centering
  \caption{Pause-scorer constants, calibrated once on known-good and known-bad phrasing
  and held fixed across systems.}
  \label{tab:pause-hyperparams}
  \small
  \begin{tabular}{llc}
    \toprule
    Stage & Parameter & Value \\
    \midrule
    \multirow{6}{*}{Silence detection}
      & Silence floor (absolute)            & $-35$ dBFS \\
      & Minimum pause duration              & 75 ms \\
      & Closure allowance, fricative onset  & 100 ms \\
      & Closure allowance, one oral stop    & 125 ms \\
      & Closure allowance, two oral stops   & 137.5 ms \\
      & Analysis hop / window               & 12.5 ms / 50 ms \\
    \midrule
    \multirow{2}{*}{Alignment}
      & Frame rate / input rate             & 50 fps / 16 kHz \\
      & Batch size                          & 8 \\
    \midrule
    \multirow{4}{*}{Attribution}
      & Boundary-snap tolerance             & 60 ms \\
      & Juncture grid                       & TLTK syllables \\
      & Word grid                           & TLTK $\cap$ newmm \\
      & Unaligned character-span threshold  & 500 ms \\
    \midrule
    \multirow{2}{*}{Artifact rejection}
      & Repeated-word slack                 & 1 syllable \\
      & Max pause share of its word span    & 0.9 \\
    \bottomrule
  \end{tabular}
\end{table}

\paragraph{Silence-detection implementation.} We exclude low-energy regions that touch a
clip boundary. For internal regions we read the phones that meet at the juncture --
the coda of the syllable on the left and the onset of the syllable on the right --
and exclude the region as a consonant closure when it is no longer than the allowance
for that context. Each allowance is the 99.5th percentile of the within-word gap
distribution measured for that context over approximately 7,200 junctures pooled across the three
systems. A glottal onset receives no allowance. The dBFS floor is absolute rather than
relative to the clip's loudness.

\paragraph{Artifact rejection.} Three conditions make a detected silence unattributable
to phrasing, and we report such silences separately rather than counting them as
placement errors. First, a repeated word: the aligner must place the single copy in the
text across both utterances, so the gap between them appears word-internal. We detect
this when the word's aligned span exceeds its syllable count plus one syllable of slack,
measured at the clip's own median onset-to-onset interval, plus the pause. Second, an
impossible attribution: a pause covering more than 0.9 of the span of the word that
supposedly contains it. Third, unverbalized input: a word abutting digits or Latin text,
whose spoken Thai form the character aligner has no vocabulary for, so the word's span
absorbs that audio. On the two reference systems, these conditions remove a small number
of unattributable silences; intra-word rates reported without them are upper bounds.

\paragraph{Mask normalization.} We keep a linguistic-mask mark only when it falls on a
word boundary. A mark written immediately before a source space is moved to the
start of the following word, avoiding an unreachable boundary. After union with the
text-derived positions described in Section~\ref{sec:pause-dataset}, the masks provide
an average of 13.8 allowed positions per sentence.

\paragraph{Alignment implementation.} We resample audio to 16 kHz, split character spans at
each silence midpoint, and snap to the nearest TLTK word boundary within 60 ms, then
onto the TLTK syllable grid, which refines the word grid. The syllable snap is what
keeps a reported juncture phonotactically possible: Thai admits no break after an
onset consonant, after a preposed vowel, or before a final consonant, so a juncture
inside a syllable can only be alignment noise. A silence inside a character span
longer than 500 ms is marked unaligned rather than misplaced; the unaligned rate is
0.0\% for all reported systems. Combining marks cannot start a pause boundary.

\paragraph{Aligner validation.} We validate three forced aligners against a
duration-predictor reference on approximately 1,500 clips synthesized by our early
duration-predictor-based Kokoro model~\citep{hexgrad2025kokoro,li2023styletts2}. The
predicted frame counts are rendered directly by the vocoder and therefore provide
token-boundary ground truth. We compare onset errors and pause decisions against this
reference.

\begin{table}[h!]
  \centering
  \caption{Forced-aligner agreement with duration-predictor ground truth. Onset errors
  are the median and 90th percentile; $r$ measures the correlation of per-voice pause
  precision with the reference, and agreement measures identical clip verdicts.}
  \label{tab:aligner-validation}
  \footnotesize
  \setlength{\tabcolsep}{4pt}
  \begin{tabular}{lccccccc}
    \toprule
    Aligner & fps & \multicolumn{2}{c}{Onset err.\ (ms)} & Pause & PPER & Per-spk. & Agr. \\
    \cmidrule(lr){3-4}
     & & p50 & p90 & prec. & & $r$ & \\
    \midrule
    Duration predictor (ref.) & --- & --- & --- & 88.1\% & 15.3\% & --- & --- \\
    \midrule
    \texttt{Typhoon ASR CTC (Whisper)} & 50 & 18.5 & 41.9 & \textbf{91.0\%} & \textbf{13.6\%} & 0.975 & \textbf{90.7\%} \\
    \texttt{wav2vec2-large-xlsr-53-th} & 50 & \textbf{13.9} & \textbf{36.3} & 89.5\% & 15.9\% & \textbf{0.984} & 88.8\% \\
    \texttt{MMS\_FA} & 50 & 24.0 & 42.7 & 68.6\% & 35.8\% & 0.724 & 74.0\% \\
    \bottomrule
  \end{tabular}
\end{table}

We use \texttt{wav2vec2-large-xlsr-53-th} for all pause measurements reported in this paper, as it is smaller and sufficiently accurate for these use cases. We also thank the Typhoon Team for generously releasing their Whisper CTC version.

\end{document}